# Neural perceptual model to global-local vision for the recognition of the logical structure of administrative documents

Boulbaba Ben Ammar

Faculty of Sciences of Sfax, Sfax University, Sfax, Tunisia

*ABSTRACT*

*This paper gives the definition of Transparent Neural Network "TNN" for the simulation of the global-local vision and its application to the segmentation of administrative document image. We have developed and have adapted a recognition method which models the contextual effects reported from studies in experimental psychology. Then, we evaluated and tested the TNN and the multi-layer perceptron "MLP", which showed its effectiveness in the field of the recognition, in order to show that the TNN is clearer for the user and more powerful on the level of the recognition. Indeed, the TNN is the only system which makes it possible to recognize the document and its structure.*

## 1. INTRODUCTION

For the moment, and although research in the field of recognition has continued for several years, the complete solution has not yet emerged. Although the machine is able to perform complex calculations and often exceeds human capacities, it remains paralyze in other areas, especially in the field of artificial intelligence. The very great quantity of documents, the variability of the continuous and the structures, the need to distinguish and to sort make automation and identification by a computer complicated. However, human is able to recognize and easily segmenting such a document only by identifying the elements which compose the logical structure. We chose to base the segmentation and the recognition on human models identification. Various approaches have been proposed to solve this problem by using a neural network [9, 10]. But none was a sufficiently simple, comprehensive and effective. In addition, most existing models do not offer a solution to the identification of a document's structure. We propose in this paper a solution to recognize the type of document and obtain its structures that compose it.

The paper is organized into three parts. The first part is devoted to the representation of an experiment made by psychologists on perception and memory in humans. Then we carry out a study of the recognition systems of the writing based on perceptual models, which use the TNN. Finally, we locate our system in this context. In the second part, we study the definition of TNN as it was defined by the authors of [3, 4]. Then we propose a study of the technique and learning algorithm used in our model. Next, we describe the structure and the topology of our model. To finish, we present an implementation of the model on an application, the administrative documents, with the TNN and the MLP, and we make a comparison between these two types of networks. In the last part, a conclusion closes the paper and gives some perspectives for future improvements.

## 2. STATE OF THE ART

### 2.1. Psycho-cognitive experiences

Psycho-cognitive experiments were performed on a number of individuals to observe the behaviour of the human being at the moment of reading [4].

1. Rumelhart and McClelland have a first experience, on a human subject, with letters isolated one after the other [5]. This subject has to press a button as soon as he sees the target letter. The measurement of response time determines the time required to recognize this letter. In a second experiment, the subject must recognize a letter in a word in order to study the effect of textual information. McClelland and Rumelhart notice that the subject recognizes faster a letter in a word, when it is shown separately. The cognitive scientists have called this phenomenon "effect of the superiority of word"[3]. It is known in recognition of the writing under the name of "contextual information".

2. The second type of experiment was carried out in this context by McClelland and Rumelhart in [6] is the study of the visual perception of a child. For this, they presented him as the representative form a typical dog. Once the child learned this form, they presented to him other incomplete forms (of dog). The child was able to supplement the presented forms. Although the forms given to the child display differences and small distortions compared to the learned typical form, this last arrived always to reproduce the general shape of the dog.

3. In a third experiment, the child observed dogs and cats. Two forms illustrate the prototypes observed. It is obvious that the prototypes of the dog and the cat are very close. Confusion between these two forms is noticed.

4. In another experiment, additional information, the name associated with each prototype, is added. After having learned three types of prototypes and their names, the child is faced to 16 examples of these three prototypes. Each example displays distortions to the level of the form as on the level of its name. No confusion was observed in children.

These experiments show that a global vision is not enough to identify the form. Confusion is detected as soon as a second form is added. The local vision is less powerful and slower if the zones of the forms to be recognized are shown separately.

### 2.2. Model reading

Psycho-cognitive experiences of Rumelhart and McClelland have observed the behaviour of the human being at the moment of reading recognition and can be inferred from the following observations [7]:

1. *Importance of lexical context:* the global vision can help to deduct local information in some distortion cases.

International Journal of Artificial Intelligence & Applications (IJAIA), Vol. 4, No. 5, September 2013

2. *Obvious characteristics:* global vision may be sufficient for the recognition of a form.
3. *Detailed analysis:* in the presence of close forms, additional information is necessary.
4. *Prototyping forms:* in order to recognize forms representing distortions, it is not necessary to learn all the possible distortions. The learning of a typical prototype can be sufficient.

On these principles, psychologists have proposed perceptual models by particular types of neural networks which were implemented by researchers in automatic reading.

**2.2.1. Interactive activation model**

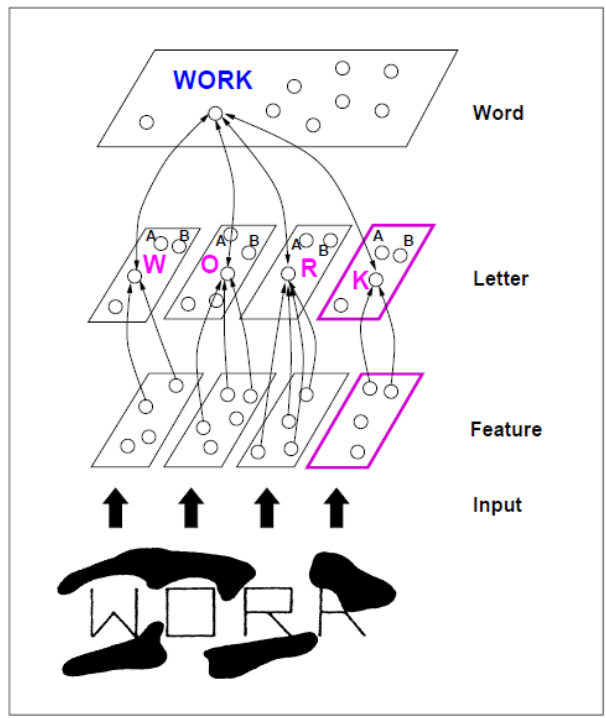

Figure 1. Interactive activation model

The model of McClelland and Rumelhart [5, 6] is based on the interactive activation through a neural network with three layers with an aim of modelling the reading of printed words composed of four letters. The layer is composed of four primitives' letters. The primitive layer consists of 16 neurons, each one corresponding to a segment having a specific orientation, called visual trait. The presence of a visual index propagates the corresponding neuron simulation through the two other layers. In back-propagation, interactions between activated neurons and the input image are made to assist the final decision. The architecture of the model of interactive verification of words reading is represented in figure 1.

**2.2.2. The verification model**

The verification model [8] of visual stimuli on the words that were activated by the latter in order to find the best candidate is based on four steps:
- Generate a set of semantically close words,

International Journal of Artificial Intelligence & Applications (IJAIA), Vol. 4, No. 5, September 2013

- Checking the validity of the semantics of words activated (stimulus) in this unit,
- Generation of a sensory unit (visual aspect)
- Checking of the visual indices in this unit. It is a question of approaching at the same time physically and semantically the words.

### 2.2.3. The two-way model

In the two-way model [2] for the recognition of words or pseudo-words, the first way proceeds by propagation of visual indices that may lead to the activation of words and pseudo-words. The second way valid the recognition of words by a phonological and/or semantics approach. Visual indices used are identical to those used in the interactive activation model of McClelland [3].

## 2.3. Perceptual recognition systems

Various perceptual systems have been investigated for the recognition of handwritten words. These systems are based on either the verification model, or on the interactive activation model, or on a combination of both.

### 2.3.1. PERCEPTRO model

PERCEPTRO model [3] is based on the model of interactive activation and the verification model. It is composed of three layers as presents in figure 2: the layer of the primitives, the layer of the letters and the layer of the words. The primitives suggested are two types: primary primitives such as the ascending ones, secondary descendants and loops and primitives such as the various forms and positions of the loops, the presence of the bar of "T", the hollows and the bumps. The primary primitives are used to initialize the system and are propagated in order to generate an initial whole of words candidates. In retro-propagation, the presence of the secondary primitives is checked following a mapping of the words candidates with the initial image. This mapping is ensured by a fuzzy function which estimates the position of the secondary primitives to check and which depends on the length of each word [3].

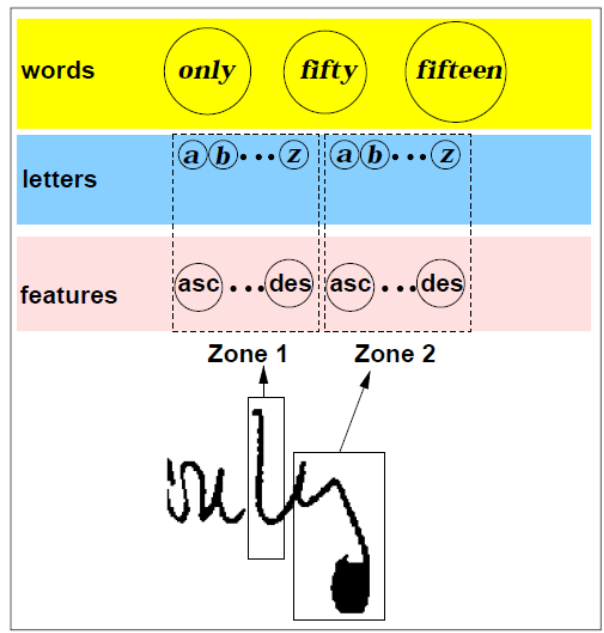

Figure 2. Architecture of PERCEPTRO system

### 2.3.2. IKRAA model

IKRAA model for the recognition of Arab words Omni-script writers is inspired by PERCEPTRO model. It is composed of four layers as presents it figure 3: the layer of the primitives, the layer of the letters, the layer of the PAW "set of related letters" and the layer of the words. As in [3], Snoussi [4], kept the same primary primitives such as the ascending ones, the descendants, the loops, the hollows and the bumps, but they are adapted to the Arab handwritten words. In retro-propagation, a standardisation based on the descriptors of Fourier (DF) is used at the local level for the identification of the letters. These descriptors are standardized and compared with those of printed letters, corresponding to the missing letters. Indeed, the standardized descriptors of Fourier are invariants compared to the position, with the size, the slope and rotation. This makes it possible to solve the problem of verification of the secondary characteristics which is an expensive and delicate step in PERCEPTRO system.

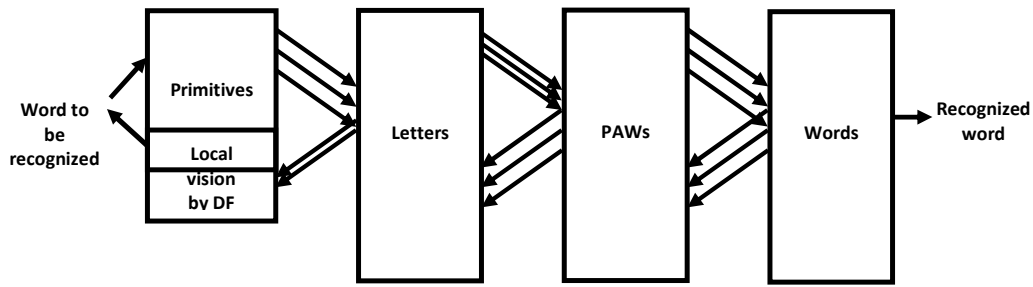

Figure 3. Architecture of IKRAA system

## 2.4. Local vision vs. global vision

### 2.4.1. Global vision

In the global vision, the document is regarded as one entity identified by direct comparison with the typical document. This approach is largely used in the recognition of the writing and of the forms because it makes it possible to avoid the problems of segmentation. The global vision is powerful in the presence of a limited vocabulary, or for the identification of forms which resemble the typical forms which were used in the learning step. It proved its effectiveness on the level as of perceptive systems in [3, 11, 12]. The psycho-cognitive experiments proved that the procedure of reading in the human starts with a global vision of the relevant characteristics.

### 2.4.2. Local vision

The document must be segmented in zones and it is the local vision which will make it possible to identify these zones, and to segment the entire document in logical structures and sub-structures. Some work shows that the recognition of the letters at the ends can be sufficient for the whole recognition of word. Moreover, sometimes an element is more easily identifiable than a document especially in the case as of administrative documents having a very complex structure. Indeed in a document, we noticed regularity in the components which build them, but also, a large variety on the level of the positioning of its components. For this reason we propose a local vision to recognize the components before identifying the document, because these components result from the elements or the zones which build them.

**2.4.3. Combination of global and local visions**

By analysing the advantages and the disadvantages of the two visions global and local, we notice a certain complementarity between them. Indeed the global vision limited the variety of the documents which can exist, because it is compared to a typical document. The local vision makes it possible to widen this variety because it is interested only in the elements or the logical structures which compose the document. Indeed in a class of documents, we generally finds the same logical structures but on different positions. Moreover this combination was proven on the level of the perceptive neural model for their cognition of Arab words Omni script writing in [4].

**2.5. Proposed approach**

PERCEPTRO and IKRAA models have the same architecture except that Snoussi added a fourth layer for related letters "PAW". This architecture was inspired by the observations of the behaviour human being at the time of reading. It was checked on the Latin and Arab words. So, we chose to adapt it to our problem. Moreover, we add a combination of global-local vision. Indeed the first procedure to recognize forms by human starts with a global vision or a vision directed by the context. Although this vision proved its effectiveness for the pattern recognition with the neural network "MLP", it appears insufficient to us for a great quantity of document because of the variability of continuous and the structures. The local vision has its turn, makes it possible to recognize the document because the logical structures can arise in various documents. It by combining the advantages of the two visions, the vision directed by the context and the vision directed by the data, which we notice a certain complementarity.

Finally, a characteristic of the natural structures consists in going from simplest to most complex. We adopted this idea to build our model in four layers, on the basis of the simple characteristics to go gradually towards the structure of the document and by highlighting the different substructures which compose it. With the new concept of transparency for the neural networks, we chose an approach which meets better research in psychology and the inherent of human characteristics.

## 3. TYPES OF NEURAL NETWORKS

We can classify neural networks in two main categories which are networks called "feed-forward" or "feedback". As the name suggests, in the feed-forward networks, information is propagated from layer to layer without turning back, unlike the feedback networks.

Neural networks can also be classified according to the type of learning, or learning rule they use. Each learning algorithm has been defined for a particular cause and therefore when we speak to learning an algorithm, a specific architecture is induced in the speech.

In this paper, we will divide the neural networks in two categories: black-boxes neural networks (MLP) and transparent neural networks (TNN).

In the following, we propose to test the learning of TNN, and to compare it with MLP.

**3.1. Learning of neural networks**

For any type of neural network, learning can be viewed as the problem of updating the weights of its connections, to achieve the task requested. Learning is the main characteristic of RNA and it can be done in different ways and according to different rules.

### 3.1.1. Types of learning

**The supervised mode:**

In this kind of learning, the network adapts by comparison between the result which it calculated and the result expected as output. Thus, the network will change until it finds the good output, i.e. which expected, corresponding to a given input.

**The unsupervised mode:**

In this case, the learning is based on a various measures of distance. The network will change according to the static regularities of the input and will establish categories, while allotting and by optimizing a value of quality, with the recognized categories.

### 3.1.2. Learning algorithm for the TNN

The problem encountered in this step is the adjustment of the weights between the links and the degrees of activation of each neuron. PERCPTREO and IKRAA models are without learning and use equations defined by McClland and Rumulhart in [6]. These equations give values to the synaptic weights calculated by statistics made on the model.

For our model, we thought about the learning rules used in the multi-layer neural networks but their use could disturb the links between the neurons. So, we proposed to divide our TNN into three monolayer networks "NN1" (see figure 4).

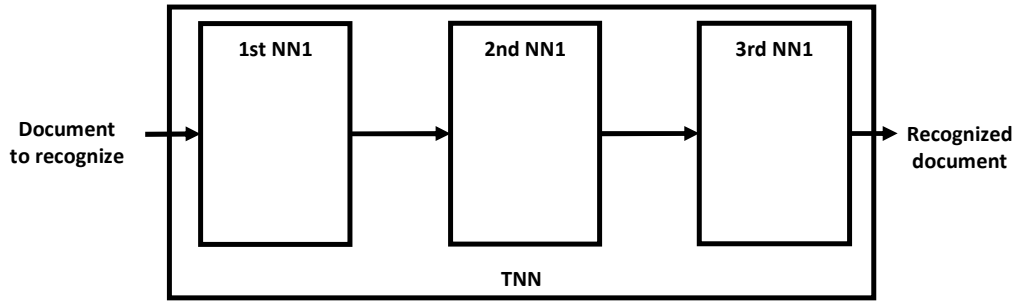

Figure 4. Division of TNN in 3 NN1s

Monolayer network is a simple network which consists only of an input layer and an output layer. It is modelled by the visual system and thus aims to pattern recognition. However, it can be used to make the classification. However, our system is composed of four layers so it took 3 NN1s for the split. Also, the first NN1 is responsible for detecting substructures in the document. The second, from the outputs of the first, detects structures in the document. The latter determines the class of the document from the outputs of the above. In this context, we conducted to the learning of each NN1 separately. Then we recovered the synaptic weights after learning of these NN1s and we injected into our TNN. Thus, our TNN has the better synaptic weights suited to the task entrusted to him.

*Step 1: calculating the error of the output layer K*
$$\delta_k = l'(S_k).E_k$$
*Step 2: calculating the $\Delta W$*
$$\Delta W_{jk} = \mu.S_j.\delta_k$$

*Step 3: modification of weight*

$$W_{jk}(t+1) = W_{jk}(t) + \Delta W_{jk}$$

With:

$E_k = desired\ output - actual\ output$

$S_k$: the actual output of the neuron
$W_{jk}$: the weight of the link between neurons j and k
$\mu$: the correction step
$l$: the threshold function
$t$: the moment t

In our case, we chose the sigmoid activation function (see figure5)

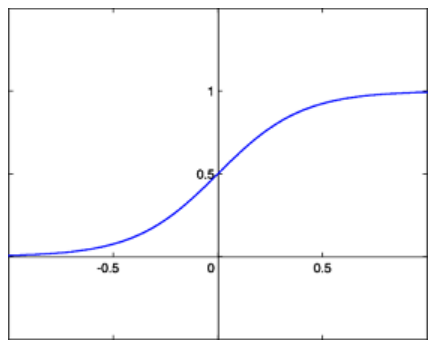

Figure 5. The graphical representation of sigmoid function

Its mathematical equation is as follows:

$$l(x) = \frac{1}{1+\exp(-x)}$$

And its derivative:

$$l'(x) = l(x).(1 - l(x))$$

We chose this function because it gives values between 0 and 1, and its derivative is calculated very quickly because it is a function of itself.

### 3.1.3. State of the neurons

In transparent neural network, each neuron is a defined concept. Whenever this concept is present, the corresponding neuron is activated. For the activation of input neurons, we propose relatively simple algorithms that extract the percentages of presence of elements in the document to recognize. In an ambiguity case, the system will go back to refine these algorithms in order to provide more precise and correct result. Each input neuron will give a value between 0 and 1, with the percentage of the presence of element in the document to recognize, for the input value of the corresponding neuron. This value can be refined in failure recognition of the document.

## 3.3. General architecture

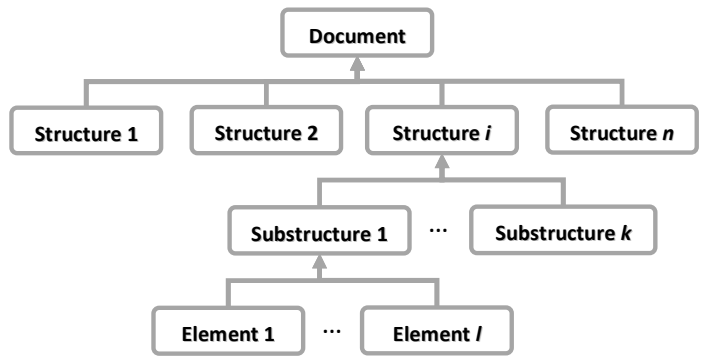

Figure 6. Hierarchy of a document

Our model is composed of four layers of organized cells in a hierarchical way and it is inspired of the structure of a document (see figure 6): a layer for the neurons elements, a layer for the neurons substructures, a layer for the neurons structures and a layer for the neurons documents. The neurons structures and the neurons substructures make it possible to extract the structure and substructures of the document. And it is this concept of transparency which enables us to deduce them. The figure 7 describes this hierarchy.

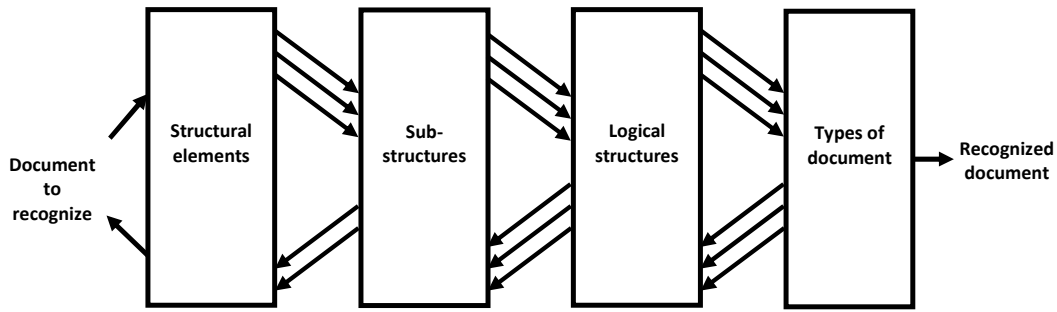

Figure 7. General architecture of TNN

## 3.4. Approach global-local vision

In section 2.4, we showed that the global vision or directed by the context is so effective and efficient if the forms of documents to recognize are close to the forms of typical documents. The document is regarded as a single entity identified. This is not the case of administrative documents in which we find a wide variety of positions in logical structures that compose them.
By cons, we detect regularity in the structures and data. We thought the vision led by the data or the local vision. But this approach is inadequate because these structures or data may exist in most documents.

For these reasons, we propose to combine these two approaches which present some complementarity. Indeed, from local to global, the elements vote for substructures, substructures vote for structures and structures vote for documents. And from global to local, if recognition of the document was a backward and extraction of structures voting for the recognized document. In

International Journal of Artificial Intelligence & Applications (IJAIA), Vol. 4, No. 5, September 2013

case of ambiguity, we return back, which will look for neurons that caused the error. In arriving at the input neurons, we propose a refinement of neural inputs corresponding to these values (see section 3.6). Once the input refined values, a new propagation will made.

To show the effectiveness of our approach and our model, we propose an application of the recognition of administrative documents types invoice, form and administrative letter. This application will be implemented with the TNN and the MLP, to make a comparison between the proposed approach and global vision and to show the effectiveness of the proposed TNN and proposed learning.

### 3.5. Application of administrative documents

#### 3.5.1. Architecture of the TNN

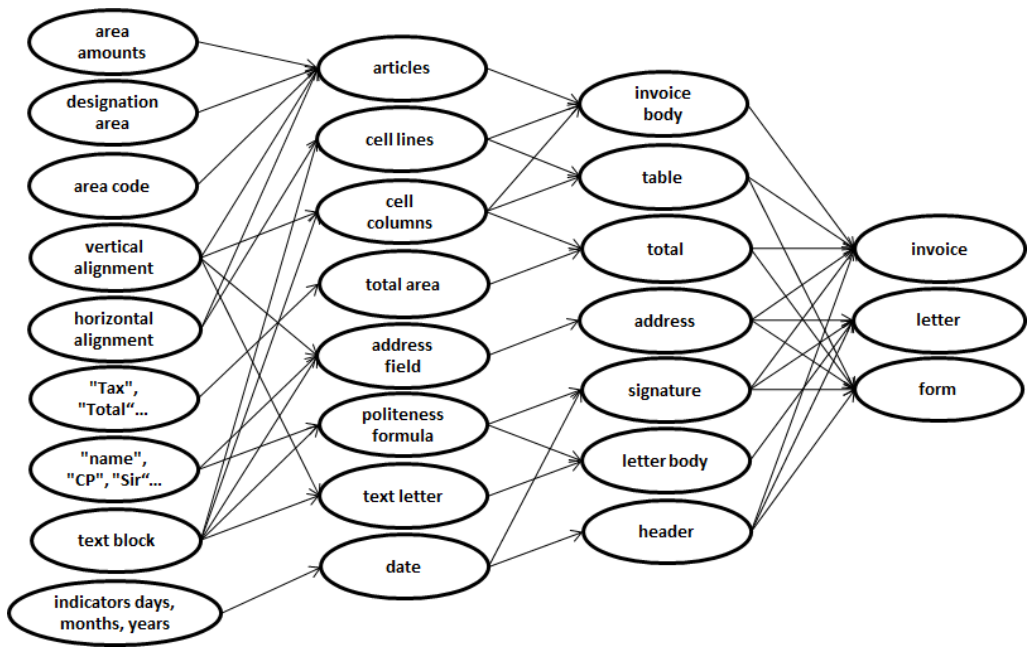

Figure 8. Architecture of neural perceptual model to global-local vision for the recognition of administrative documents and extraction of its logical structure

The figure 8 shows the various neurons, as well as the links between these neurons and the various layers which build the model (the links and the number of neurons were deducted from the hierarchy of each type of documents). As we chose the human model perception, the transparent neural network is adapted to our needs. With the concept of transparency, we chose an approach which gives the better answers to research in psychology. Our model comprises four layers:

- The last layer comprises the various documents which wishes to study. To validate our system, we chose three types of documents: invoice, form and letter. So, the number of neurons of the output layer is three.
- The third layer is made up of as many neurons as structures which compose the various documents, from where links between these two layers. The concept of transparency will make it possible to recognize all the structures which build the document result.

International Journal of Artificial Intelligence & Applications (IJAIA), Vol. 4, No. 5, September 2013

- The neurons of the second layer correspond to substructures which compose the various structures of the previous layer.
- Finally the input layer comprises the elements or the characteristics which can find in the different substructures.

Our choice consists on starting with the simple elements to go gradually towards the more complex structure of the document by highlighting the different substructures which compose it.

### 3.5.2. Architecture of the MLP

The figure 9 represents a multi-layer neural network or MLP for the recognition of administrative documents.

The multi-layer neural network is composed of four layers. The first and the last layer are identical to those proposed in the TNN. The two other layers were private of their conceptual role, and became hidden.

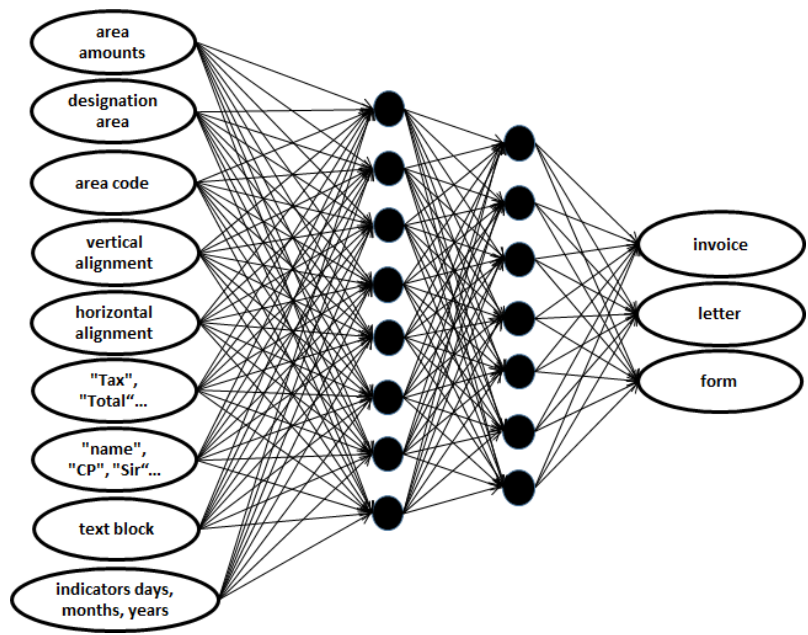

Figure 9. Architecture of an MLP for the recognition of administrative document

The MLP makes it possible to recognize the document but not its structure. It uses a global vision or a cognition directed by the context. The algorithm of the retro-propagation of the gradient was used to update of the synaptic weights of the multi-layer neural network.

### 3.6. Description of activation

A neuron can be either passive or active. The value of its internal energy varies between 0 and 1 (see figure 5). When a neuron is stimulated, its degree of activation grows and little to influence its neighbours. They are the neurons of the higher level, in the case of the propagation, or inferior level, in the case of the retro-propagation, to which it is connected.

### 3.6.1. Input values

The input value of each neuron of the first layer can be obtained by one or more algorithms. These algorithms are refined gradually in order to obtain an increasingly precise value. At the beginning of each recognition, all the neurons of the first layer use their first algorithm to obtain an input value between 0 and 1. In the ambiguity event, the network will return to the neurons elements, which were classified by the network as the causes of the error. These neurons will activate their second algorithm of extraction of the characteristics in order to obtain values of more precise entries. If necessary, a third passage will be carried out. The order of the algorithms is selected so that the first is the least expensive in term of computing power and consequently the least precise. The last algorithm is most precise and most expensive. The specific characteristics to each algorithm are detailed by neuron as follows:

*Total area:*
1. This zone is characterized by the presence of only numbers
2. Two alignment conditions must be met: vertical and horizontal
3. It is necessary to validate the formula "quantity * price = amount" between three columns of numbers

*Zone designation*
1. This zone is characterized by the presence of alphanumeric with a majority of alphabetic
2. These should be aligned vertically
3. The designation area is between the area code and the amount area

*Area code*
1. This zone is characterized by the presence of alphabetic with a full majority
2. A condition in this zone is a vertical alignment between the elements
3. This is generally located leftmost

*Vertical alignment*
1. It is detected by the left justify
2. It can be refined taking into account the right justify

*Horizontal alignment*
1. It is characterized by the presence of regular spacing between the items

*Grouping keywords "Tax", "Total"*
1. A condition of this area is the presence of "VAT" and "Total" chains
2. The presence of other keywords can validate this area: tax, VAT, amount, net pay...

*Grouping keywords "name", "CP", "Sir"...*
1. A condition of this area is the presence of the chains "Mr." , "Mrs." , "Name", "Name", "postal Code", "Town and Country", "street", "codex", "BP" ...
2. Positions are analysed and the values associated with these channels

*Text block*
1. This zone is characterized by the presence of only alphabetic
2. A paragraph and its left justify may be characteristics of this area

*Indicators days, months, years*
1. A condition of this area is the presence of a string "dd/mm/yyyy ' dd-mm-yyyy", "dd/mm/yy", "dd-mm-yy" ...
2. Function date recognition that details and analyses are used in the various fields

Each characteristic makes it possible to describe a calculating algorithm for the value of activation of neuron. These values were simulated for our application. With regard to the learning, the values are selected by integrating all the characteristics.

### 3.6.2. Activation function

A neural network is a directed and balanced graph. The nodes of this graph are the formal neurons which are simple automats equipped with an internal state, this state represents by the activation which they influence the other neurons of the network. This activity is propagated in the graph along balanced arcs, they are the synaptic links. The weighting of a synaptic link is called "synaptic weight". The rule which determines the activation of a neuron according to the influence of the other neurons of network is called "function of activation".

*Function of activation*

The variables of this function are called "inputs" of the neuron and the value of the function is called "output". Thus we can consider a neuron as a mathematical operator which one can calculate the digital value by a computer program. In our case, we chose the following function:

$$a_j = \sum (W_{ij} \cdot e_i) - \theta$$

With:
$a_j$ : the activation of neuron $j$
$W_{ij}$ : the weight of the connection between neuron $i$ and neuron $j$
$e_i$ : the input of neuron $j$, or the output of neuron $i$
$\theta$ : the threshold of activation
Neuron $J$: it is a neuron of the previous layer andwhich has a link with the neuron

*Synapse weight:*

In the perceptive models of Côté [3] and Snoussi [4], the weights of connection do not require a learning step. Indeed, these weights are estimated according to the information represented in the neurons compared to the sub-base, and as these weights are estimates, the probability of the error is stronger. This is why we thought of al step (see section 3.1).

### 3.7. Results

#### 3.7.1. Manual extraction of elements

We evaluated the TNN and the MLP without integration of the extraction step of the characteristics. The characteristics are visually described by the human and introduced in the two recognition systems. It is necessary to take into account in the continuation of this experimentation that the recognition rate depends on the automatic quality of extraction of the elements.

International Journal of Artificial Intelligence & Applications (IJAIA), Vol. 4, No. 5, September 2013### 3.7.2. Performances of the TNN

*Document Recognition*

To document recognition, the system returns back to change the input values of the neurons of the first layer that caused the failure. Table 1 presents the results of this experiment.

Table 1. Recognition rate of documents relating to the TNN

| types of document | number of apprentices document | number of documents tested | number of recognized documents | recognition rate |
|---|---|---|---|---|
| invoices | 40 | 120 | 118 | 98.33% |
| forms | 36 | 90 | 87 | 96.66% |
| letters | 26 | 40 | 39 | 97.5% |
| all documents | 102 | 250 | 244 | 97.6% |

*Extracting document structure*

The performance evaluation is carried out as well for the extraction of the structures as for their cognition of the documents. Table 2 presents the various results. We noticed that even in the event of failure at the time of the recognition of a document, our system allows the extraction of its structure. In the same way, the system is able to recognize the type of a document without the identification of all the structures which compose it.

Table 2. Recognition rate of document structures

| types of structures | number of structures tested | number of recognized structures | recognition rate |
|---|---|---|---|
| *invoice body* | 120 | 118 | 98.33% |
| *table* | 179 | 177 | 98.88% |
| *total* | 190 | 190 | 100% |
| *address* | 250 | 250 | 100% |
| *signature* | 250 | 227 | 90.8% |
| *letter body* | 40 | 39 | 97.5% |
| *header* | 250 | 247 | 98.8% |
| *all structures* | 1320 | 1248 | 94.54% |

The class of a document cannot be recognized because the absence of the structures which composes it (the absence of the characteristics of the input neurons), or the presence of other unusual structures (text in an invoice) or its resemblance to another class.

### 3.7.3. Performances of the MLP

To show the effectiveness of our system, experimentation was made with a MLP. Table 3 shows the results of this experimentation.

102102

Table 3. Recognition rate of documents relating to the MLP

| types of document | number of apprentices document | number of documents tested | number of recognized documents | recognition rate |
|---|---|---|---|---|
| invoices | 128 | 120 | 108 | 90% |
| forms | 128 | 90 | 87 | 96.66% |
| letters | 128 | 40 | 35 | 87.5% |
| all documents | 384 | 250 | 230 | 92% |

The documents tested with the TNN and the MLP are the same ones. But the MLP requires more samples to carry out the learning step. Some of these samples were re-used for the recognition.

### 3.8. Comparison

#### 3.8.1. On the level of the learning

To have a finer control on the system, it is necessary to act on the values of the weights of connections. However, the choice of a set of correct synaptic weights is a difficult problem. This is why we propose a learning algorithm which makes it possible to adjust them automatically, contrary to PERCEPTRO and IKRAA systems.

The TNN used a learning algorithm as it was described in the section 3.1. It was carried out with 40 invoices, 36 forms and 26 letters for the system to become stable. On the other hand, the learning of the weights in the MLP required 128 invoices, 128 forms and 128 letters. Moreover, the learning step for the MLP generated 10 times more retro-propagation to stabilize the system than the learning for the TNN.

#### 3.8.2. On the level of the recognition

By comparing table 1 and table 3, we notice that the two networks allow the recognition of documents with satisfactory rates (97.6% for the TNN and 92% for the MLP). But we notice that the network of black-box neurons requires a high duration of learning as well as a broad database.

In addition, the TNN makes it possible to classify the document and to know its structure whereas the MLP authorizes only the ranking.

## 4. CONCLUSIONS

This work falls under the will to bring to sciences of cognition a new neuronal perceptive model to global-local vision for the recognition of document and its logical structure.
We began our research with a bibliographical study of the existing systems and methods applied for the recognition. We were interested in the perceptive systems inspired by the models of recognition defined by the psychologists.

Various perceptive systems were based on the modelling of the humans reading, in particular PERCEPTRO and IKRAA systems which were developed for the recognition of the Latin and Arab manuscript writing.
We proposed a perceptive neural model TNN, allowing to simulate the global-local vision for the recognition and the segmentation of the image of a document. We studied the architecture of the TNN and its topology deduced from the hierarchy of a document. We also introduced a learning algorithm on this kind of network. Indeed, a TNN can be regarded as several NN1s (monolayer neural network) put in cascade. With each one of these NN1s is entrusted the spot to admit a particular level of the architecture of the object to be recognized. In our case, the TNN charged to recognize the class of the documents and their structure is composed of 3 NN1s. The first NN1 is charged to detect substructures of the document. The second, starting from the outputs of the first, detects the structures present in the document. The last determines the class of the document starting from the outputs of previous. Accordingly, we carried out the learning of each NN1s separately. Then we recovered the synaptic weights of these NN1s after learning and injected them into our TNN. Thus, our TNN has better synaptic weights adapted to the task which is entrusted to him.

We evaluated the TNN and the MLP on an application concerning the administrative documents and showed that the TNN is more effective than the MLP. Indeed, a MLP simulates only one global vision and thus allows only the recognition of the class of the documents. Contrary, our TNN simulates a global-local vision allowing not only the recognition of the class of the documents but also their structures. Following the recognition of the class of a document, it is possible to extract the structure of the document using the votes from the various neurons: passage of global (class of the document) towards the local (structures of the document). This same passage, following an ambiguous recognition, is used to identify the neurons of the layer of inputs which caused an error in order to refine their values of entries.

We evaluated the TNN and the MLP without integration of the extraction step of the characteristics. Those are visually described by the human and introduced in the two systems of recognition. We propose to automate this step of extraction of the characteristics or elements of entries and to assess the two systems again. The second prospect considered is the generation of the application of this perceptive neural system to global-local vision for the recognition of any type of documents and the extraction of their logical structure. Finally the last prospect is generalized our idea of learning on all types of TNN.

## ACKNOWLEDGEMENTS

I wish to thank Mr. Abdel Belaïd and all the READ team for their greeting throughout my internship in LORIA, France.

**Author**

Boulbaba Ben Ammar was born in 1981. He earned PhD degree in computer science in 2012. Actually, he is an assistant professor in computer science in Sfax University.

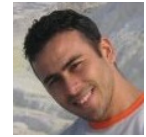